\documentclass{article}
\usepackage{spconf,graphicx}
\usepackage{indentfirst}
\usepackage{amssymb}
\usepackage{amsmath}
\usepackage{color}

\usepackage{multirow}
\usepackage{booktabs}
\usepackage{caption}
\usepackage{overpic}
\usepackage{comment}
\usepackage[normalem]{ulem}
\usepackage{subcaption}
\useunder{\uline}{\ul}{}

\usepackage{hyperref}
\usepackage[table,xcdraw]{xcolor}

\newcommand{\Fusion}{Fusion-based\cite{Fusion_based}}
\newcommand{\Retinex}{Retinex-based\cite{Retinex_based}}
\newcommand{\MLLE}{MLLE\cite{MLLE}}
\newcommand{\UDCP}{UDCP\cite{UDCP}}
\newcommand{\RED}{RED\cite{RED}}
\newcommand{\UWCNN}{UWCNN\cite{UWCNN}}
\newcommand{\WaterNet}{Water-Net\cite{WaterNet_UIEBD}}
\newcommand{\FUnIE}{FUnIE\cite{FUnIE}}
\newcommand{\Ucolor}{Ucolor\cite{Ucolor}}
\newcommand{\Ushape}{U-shape\cite{u-shape}}

\newcommand{\UDAformer}{UDAformer\cite{UDAformer}}
\newcommand{\HFM}{HFM\cite{HFM}}
\newcommand{\DMwater}{DM-water\cite{DM-water}}
\newcommand{\WFDiff}{WF-Diff\cite{WF-Diff}}
\newcommand{\GuidedHybSensUIR}{GuidedHybSensUIR\cite{GuidedHybSensUIR}}

\graphicspath{{Fig/}}

\hypersetup{
    colorlinks=true, %
    linkcolor=red, %
    citecolor=green, %
    urlcolor=blue %
}

\DeclareCaptionLabelFormat{myformat_table}{\textbf{Table. #2.}}
\DeclareCaptionLabelFormat{myformat_figure}{\textbf{Fig. #2.}}
 \pdfoutput=1

\begin{document}
\title{Underwater Image Enhancement via Dehazing and Color Restoration}
\name{Chengqin Wu$^{1}$\quad Shuai Yu$^{2}$\quad Tuyan Luo$^{3}$\quad Qiuhua Rao$^{3}$\quad Qingson Hu$^{1}$\quad Jingxiang Xu$^{1}$\quad Lijun Zhang$^{1}$}

\address{$^{1}$ Shanghai Ocean University\\
$^2$ Chinese Academy of Sciences\\
$^3$ Fujian Academy of Agricultural Sciences}
\maketitle

\begin{abstract}
Underwater visual imaging is crucial for marine engineering, but it suffers from low contrast, blurriness, and color degradation, which hinders downstream analysis. Existing underwater image enhancement methods often treat the haze and color cast as a unified degradation process, neglecting their inherent independence while overlooking their synergistic relationship. To overcome this limitation, we propose a Vision Transformer (ViT)-based network (referred to as WaterFormer) to improve underwater image quality. WaterFormer contains three major components: a dehazing block (DehazeFormer Block) to capture the self-correlated haze features and extract deep-level features, a Color Restoration Block (CRB) to capture self-correlated color cast features, and a Channel Fusion Block (CFB) that dynamically integrates these decoupled features to achieve comprehensive enhancement. 
To ensure authenticity, a soft reconstruction layer based on the underwater imaging physics model is included. Further, a Chromatic Consistency Loss and Sobel Color Loss are designed to respectively preserve color fidelity and enhance structural details during network training. 
Comprehensive experimental results demonstrate that WaterFormer outperforms other state-of-the-art methods in enhancing underwater images.
\end{abstract}
\begin{keywords}
Degradation process, ViT model, Underwater Image Enhancement
\end{keywords}
%
\section{Introduction}
\label{sec:intro}
The ocean holds vast natural resources, so deep-sea exploration and development are crucial. Underwater imaging serves as a vital carrier of marine information and image quality greatly influences marine engineering. However, with its floating particles, light scattering and selective light absorption, the deep ocean environment introduces noise, blur, and color distortion~\cite{schettini2010underwater}, which have detrimental effects on underwater detection, underwater visual perception, and other underwater tasks~\cite{underwater_task, underwater_task2}. Underwater Image Enhancement (UIE) aims to mitigate these issues, restore the true colors, enhance the details, and consequently increase the reliability and usability of the data. Therefore, research on UIE algorithms is necessary, urgent, and of great significance.

In the early days, specialized hardware~\cite{he2004divergent}, multiple different images~\cite{treibitz2012turbid} or special imaging methods~\cite{nascimento2009stereo} were used to improve the quality of underwater images. Although these methods were effective in certain scenarios, their expensive hardware and complex conditions limited their widespread use. To achieve convenience and universality, increasingly many image processing and computer vision techniques are used to address the challenges of underwater imaging. These techniques can be broadly categorized into non-physical model-based methods, physical model-based methods, and data-driven methods.
Non-physical model-based methods do not rely on underwater optical propagation models. They adjust in the spatial or transform domain to improve the visual quality of the image~\cite{underwater2010,underwater2019}. Non-physical model-based methods are user-friendly but may exacerbate color distortion or introduce unnatural artifacts due to their reliance on statistical/empirical adjustments, which ignore the physics of wavelength-selective light attenuation underwater. Moreover, most approaches decouple dehazing and color correction into separate stages, causing iterative error accumulation: information lost in earlier stages cannot be retrieved later.
Physical model-based methods enhance underwater images by analyzing the principles of light propagation, formulating assumptions, and constructing mathematical models to simulate underwater processes. For example, these approaches often estimate critical parameters such as light scattering coefficients~\cite{underwater2007} and medium transmission rates~\cite{underwater2011,transmission1}. Using the analogous modeling frameworks shared between underwater imaging and atmospheric haze imaging, these methods achieve simultaneous color restoration and dehazing with high fidelity. However, their effectiveness diminishes in complex underwater environments due to the overdependence of simplified assumptions, such as uniform illumination and static water conditions, which fail to account for real-world challenges such as artificial directional lighting, dynamic turbulence, and suspended particulate matter. Additionally, the interdependent nature of color correction and dehazing introduces mutual performance degradation when ideal conditions for either task (e.g., accurate parameter estimation or stable scene properties) are not met.
In recent years, data-driven techniques have gained significant momentum and more researchers have begun to use deep learning to enhance underwater images~\cite{UWCNN, WaterNet_UIEBD}. With their large parameter space and powerful learning capabilities, deep learning models often outperform traditional methods when ample datasets are available. However, data-driven methods face three challenges. First, most data-driven methods face a trade-off between performance and the number of parameters. Second, most of them only focus on deep feature extraction or color-based guidance, neglecting haze features' importance, which limits their UIE performance in dehazing and edge detail retention. Third, they often cannot guarantee the authenticity of the enhanced data due to the lack of physical model guidance.

Hence, we devised WaterFormer, which is a lightweight but high-performance network that considers the haze and color cast as inherent primary degradation features of the underwater environment to restore images. Specifically, WaterFormer uses the DehazeFormer Block proposed in~\cite{DehazeFormer} to achieve self-attention for the haze features and extract deep features. Using the channel-wise self-attention matrix transformation, our Color Restoration Block (CRB) performs self-attention on color cast features to help the network infer results with a more realistic color distribution. We also introduced the Channel Fusion Block (CFB), which synthesized the global information from different features and refined it to generate more effective fusion features. 
Considering the constraints of the underwater imaging physical model, we incorporated an underwater soft reconstruction layer at the end of the network to enhance the fidelity of the results. Furthermore, we introduced the Chromatic Consistency Loss and Sobel Color Loss to train the network to maintain chromatic consistency, preserve fine color details, and enhance image quality and generalization across various datasets. 

Overall, the main contributions of this work are as follows:
\begin{itemize}
  \item A novel end-to-end network is proposed to enhance underwater images. Our approach uses the DehazeFormer Block, successfully transforms haze features, and improves the image deblurring capability. It introduces an underwater soft reconstruction layer to ensure the authenticity of the results.
  \item We integrate the CRB and CFB with an encoder-decoder structure based on the global information between channels to generate a more realistic color distribution and accurately fuse features from different stages.
  \item We propose the Chromatic Consistency Loss and Sobel Color Loss to improve the training process of the network and enhance its generalization ability.
\end{itemize}

\section{Related Work}
\label{sec:relatedwork}

\subsection{Traditional Methods}
Traditional methods can be broadly categorized into methods based on non-physical models and those based on physical models. Non-physical model-based methods do not involve modeling the underwater optical transmission process. Instead, they adjust the pixels of an image in the spatial domain or transform domain using the characteristic statistics and visual perceptual analysis. These methods include contrast enhancement~\cite{MLLE}, histogram equalization~\cite{hist2}, wavelet transform~\cite{wavelet}, fusion-based methods~\cite{Fusion_based}, retinex-based methods~\cite{Retinex_based}, and others.
For example, Hitam et al.~\cite{hist2} applied the mixture Contrast Limited Adaptive Histogram Equalization to underwater images in the RGB and HSV color spaces. Singh et al.~\cite{wavelet} performed a discrete wavelet transform on images and estimated the color correction factors that were associated with specific color casts to adjust the input image pixels. Ancuti et al.~\cite{Fusion_based} separately conducted color correction and contrast enhancement on underwater images and subsequently fused the two versions based on automatically calculated weights.
Although most non-physical model-based methods are simple to implement and computationally efficient, they lack prior knowledge of the optical propagation, which can result in oversaturated or undersaturated color distributions.

Physical model-based methods are commonly used in UIE. They involve mathematical modeling of the underwater degradation process, estimating the parameters of the degradation model, and ultimately recovering clear images. These methods address various aspects of underwater image degradation, such as estimating the parameters related to the light scattering~\cite{underwater2007} and medium transmission~\cite{underwater2011,transmission2}.
For example, Hou et al.~\cite{underwater2007} used optical response functions, including Point Spread Function (PSF) and Modulation Transfer Function (MTF), and deconvolution techniques for image restoration. Ke et al.~\cite{transmission1} considered the color, saturation, and detail information in images to construct scene depth maps and edge maps to estimate the medium transmission rates. Galdran et al.~\cite{RED} adjusted the Dark Channel Prior (DCP) method by incorporating red channel priors to estimate the medium transmission map for underwater scenes. 
Physical model-based methods are based on prior assumptions and offer high fidelity in image restoration. However, their applicability is often limited to specific underwater degradation phenomena, so they have lower generality.

\subsection{Data-driven Methods}
The use of deep learning techniques in UIE tasks has been a major trend in recent years. These approaches mainly follow two paradigms: the encoding-decoding paradigm and the generative paradigm.
On the one hand, methods based on the encoding-decoding paradigm typically encode images into a latent space and continuously learn low-level features and deep representations of images to reconstruct the final results through a decoder. Encoding-decoding methods often use global residual connections or gate fusion mechanisms to facilitate network learning. Li et al.~\cite{WaterNet_UIEBD} introduced a gate fusion network called Water-Net, which enhanced images by fusing the original image with white-balanced, histogram-equalized, and gamma-corrected versions. Water-Net can adapt to different underwater environments and enhance the image. Xue et al.~\cite{MBANet} analyzed underwater degradation factors from two perspectives: color distortions and veil effects. They developed the Multi-Branch Multi-Variable Network (MBANet) to restore underwater images. Li et al.~\cite{Ucolor} proposed an enhancement network called Ucolor, which operated in the HSV, RGB, and Lab color spaces to encode images and decode image features with a medium transmission map. Therefore, Ucolor can learn and selectively enhance different visual representations of images. 

On the other hand, Generative Adversarial Networks (GANs) and Diffusion Models constitute two pivotal methodologies within the generative paradigm. GANs comprise two components: a generator that synthesizes samples from random noise, and a discriminator that differentiates authentic samples from synthetic ones. Through adversarial optimization, both components co-evolve until the generator attains photorealistic synthesis capability.  
Diffusion models involve a forward diffusion process and a reverse generation stage. During the forward process, incremental Gaussian noise is systematically introduced to an image until its statistical properties converge to a standard normal distribution, concurrently training a Unet~\cite{Unet} to estimate the noise profile. The reverse phase leverages the trained Unet to progressively refine the corrupted latent representation into a coherent image through iterative denoising operations. Islam et al.~\cite{FUnIE} designed an adversarial network called FUnIE based on conditional generation, where the generator was constructed using Unet architecture and the discriminator was a Markovian discriminator. Cong et al.~\cite{PUGAN} proposed PUGAN, which guided the generator to produce more interpretable results by estimating medium transmission parameters. Tang et al.~\cite{DM-water} developed a diffusion model-based UIE framework, employing piecewise sampling and evolutionary algorithm search to optimize generation efficiency. Zhao et al.~\cite{WF-Diff} innovatively integrated frequency domain representations into the diffusion architecture, where a Wavelet-based Fourier interaction network and Frequency Residual Diffusion Adjustment Module collectively demonstrated the critical role of spectral characteristics in UIE tasks.

Most studies on UIE tasks are based on Convolutional Neural Networks (CNNs) because they can effectively capture local features in images through convolutional and pooling layers. In recent years, the ViT~\cite{ViT} was proposed, and it can better capture both local and global perspectives of images and gain more similarity between features than CNNs. As a result, some researchers began using ViT to handle UIE tasks. For example, Peng et al.~\cite{u-shape} used the ViT architecture to build a generative adversarial paradigm network called U-shape. Shen et al.~\cite{UDAformer} constructed a ViT network called UDAformer that included a spatial attention mechanism and a channel attention mechanism in an encoder-decoder paradigm.

Although many data-driven algorithms exhibit excellent generalization, they often grapple with a trade-off between parameter count and performance. These methods overlook the interconnected but independent nature of the haze and color distortion phenomena. Treating these features as a singular degradation process neglects the interplay between haze removal and color restoration. Furthermore, existing approaches lack tailored loss functions for UIE, which decreases the learning capacity of the network. Hence, we developed a lightweight, high-performance network for haze and color distortion features with two proposed loss functions to enhance the underwater image features and optimize the learning capability of the network.

\section{Proposed Method}
\label{sec:proposedmethod}

\begin{figure*}[ht]
    \centering
    \includegraphics[width=\textwidth]{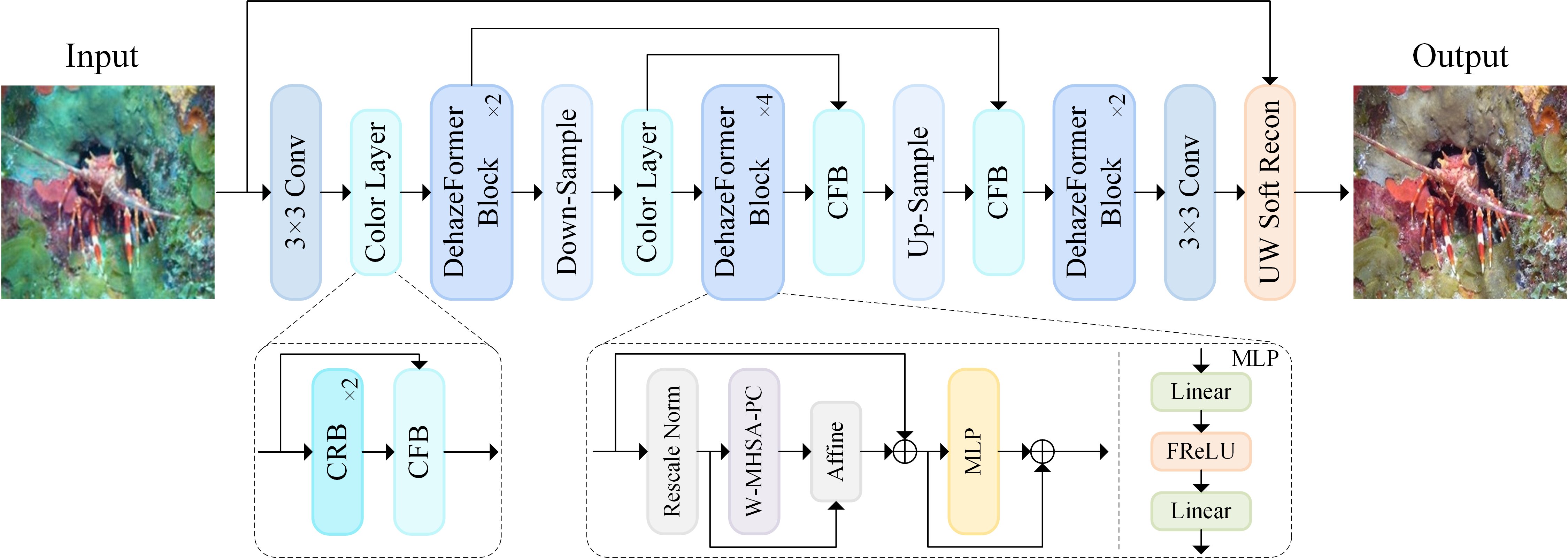}
    \caption{Overall structure of WaterFormer, which includes the DehazeFormer Block as the core component for feature extraction and transformation in the network. It's ability to aggregate local information is enhanced through the incorporation of the FReLU in the MLP. Additionally, the proposed CRB and CFB are employed to improve the color features of underwater images and facilitate the integration of features across different branches. WaterFormer is built as a three-stage, enhanced Unet network, with Down-Sample and Up-Sample stages implemented using convolution and pixelshuffle, respectively. Input/output images are from the EUVP dataset\cite{FUnIE}.}\label{fig:WaterFormer}
\end{figure*}

\begin{table*}[ht]
\centering
\scalebox{0.9}
{
\begin{tabular}{c||ccccc}
\hline
Block Type              & Num. of Blocks & Embedding Dims   & Num. of Heads & MLP Ratio     & Act. Function of MLP \\ \hline
DehazeFormer Block      & {[}2, 4, 2{]}  & {[}24, 48, 24{]} & {[}3, 6, 3{]} & {[}3, 4, 3{]} & FReLU                \\
Color Restoration Block & {[}2, 2{]}     & {[}24, 48{]}     & {[}3, 6{]}    & {[}3, 4{]}    & FReLU                \\ \hline
\end{tabular}
}
\caption{Detailed structure of the Transformer block.} \label{tab:DetailedTransformerBlock}
\end{table*}

\subsection{Overall framework and hierarchical structure}
The proposed WaterFormer adopts the U-shaped\cite{Unet} encoder-decoder architecture as the backbone.
In the encoder part, the network uses the DehazeFormer Block\cite{DehazeFormer} with self-correlation to refine the haze features and deep features. The Color Restoration Block (CRB) was designed to enhance the color characteristics of the feature maps. In the decoder part, in addition to refined features, we used the CFB for skip connections to efficiently fuse different features at different stages. Finally, to improve the realism of the estimated image, we incorporated an underwater soft reconstruction layer into WaterFormer at the end, which innovated based on the soft reconstruction layer in~\cite{DehazeFormer} and replaced the original global residual learning. Specifically, the underwater image formation process in~\cite{underwaterFormula} is as follows:
\begin{equation}\label{equ:WaterImageFormula}
{{U}_{\lambda }}(x)={{I}_{\lambda }}(x)\cdot {{T}_{\lambda }}(x)+{{A}_{\lambda }}\cdot (1-{{T}_{\lambda }}(x))
\end{equation}

\noindent where ${{U}_{\lambda }}(x)$ is the underwater image captured by the camera; ${{I}_{\lambda }}(x)$ is the clear latent image; ${{T}_{\lambda }}(x)$ is the transmission map; ${A}_{\lambda }$ is the global background light; ${\lambda }$ is the wavelength of light in the R, G, B channels. According to Eq.~\ref{equ:WaterImageFormula}, the recovery process of the underwater image can be described as:
\begin{equation}\label{equ:GTImageFormula}
{{I}_{\lambda }}(x)={{U}_{\lambda }}(x)\cdot (\frac{1}{{{T}_{\lambda }}(x)}-1)+{{A}_{\lambda }}\cdot (1-\frac{1}{{{T}_{\lambda }}(x)})+{{U}_{\lambda }}(x)
\end{equation}

We set $K_{\lambda }(x)=\frac{1}{{{T}_{\lambda }}(x)}-1$ and $B_{\lambda }(x)={{A}_{\lambda }}\cdot (\frac{1}{{{T}_{\lambda }}(x)}-1)$ to obtain the simplified Eq.~\ref{equ:WaterFormerImageFormula}:
\begin{equation}\label{equ:WaterFormerImageFormula}
{{I}_{\lambda }}(x)={{U}_{\lambda }}(x)\cdot {{K}_{\lambda }}(x)-{{B}_{\lambda }}(x)+{{U}_{\lambda }}(x)
\end{equation}

This equation can be introduced as a prior into our network. Specifically, we drove the network to predict the variable $O\in {{\mathbb{R}}^{h\times w\times 6}}$ and decomposed the variable into $K_{\lambda }(x)\in {{\mathbb{R}}^{h\times w\times 3}}$ and $B_{\lambda }(x)\in {{\mathbb{R}}^{h\times w\times 3}}$. Eq.~\ref{equ:WaterFormerImageFormula} is referenced to reconstruct the final ground truth estimation image.

To reduce the complexity and computational cost of the network, we removed a significant number of convolutional layers from the original DehazeFormer Block structure and only retained the convolution operations related to the self-attention and multilayer perceptron (MLP). To compensate for the limited utilization of local information due to the decrease in number of convolutional layers, we replaced the activation function in the MLP with a more locally enhanced FReLU~\cite{FReLU} and designed WaterFormer as a three-stage Unet-based network. If there are more complex underwater environments and less demand for network speed performance, the number of encoding-decoding stages can be increased to enhance the network performance. Fig.~\ref{fig:WaterFormer} and Tab.~\ref{tab:DetailedTransformerBlock} present the detailed structure of WaterFormer.

\subsubsection{Color Restoration Block (CRB)}
\begin{figure}[htb]
    \centering
    \includegraphics[width=\columnwidth]{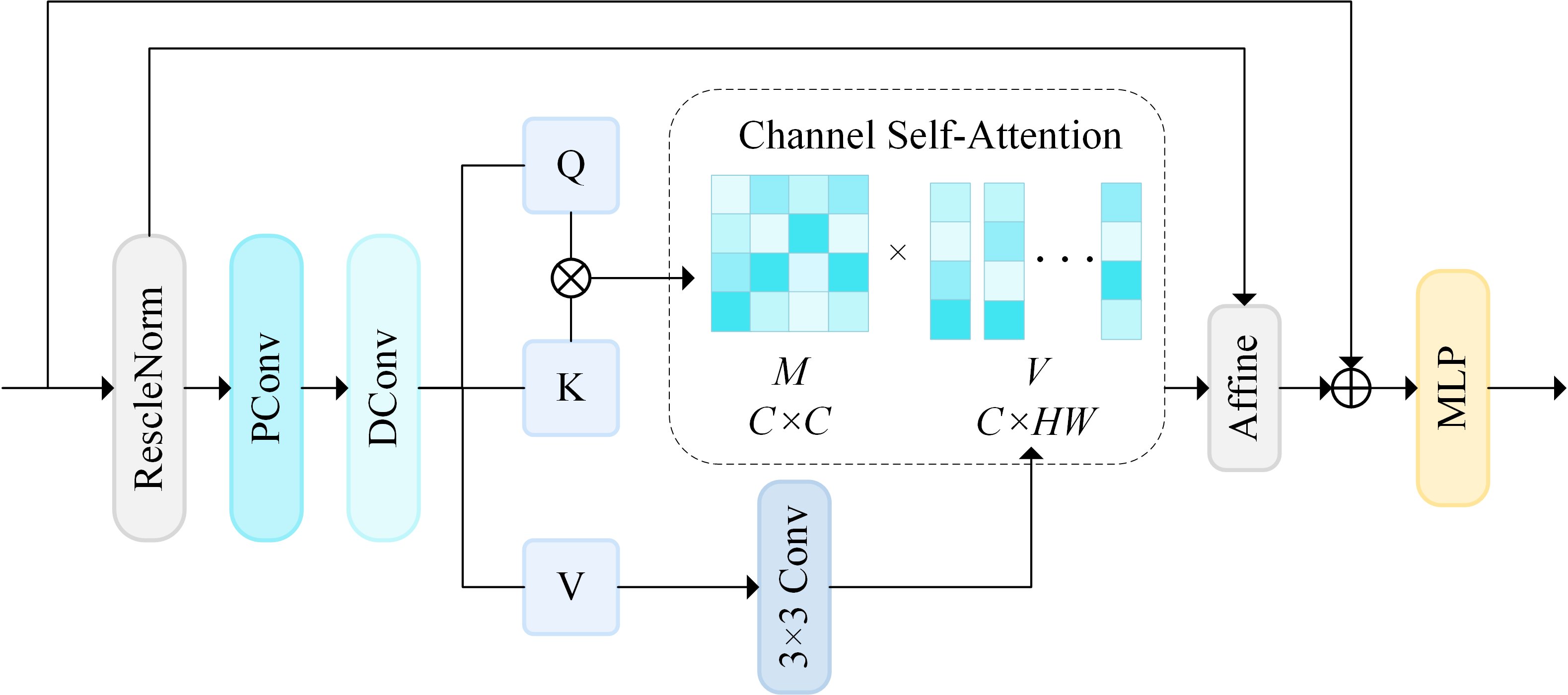}
    \caption{Overall structure of the CRB. Through channel-wise self-attention, WaterFormer can better self-transform and enhance color features in images.}\label{fig:CRB}
\end{figure}

Due to the absorption and scattering of light in water, underwater images tend to appear bluish or greenish compared to land images, which is a channel-level per-pixel interference. Therefore, leveraging the global information among the channels for color balancing is an important task in UIE. Inspired by the matrix transformations in ~\cite{Restormer,DNF}, we designed the CRB to perform matrix transformations on the pixel-wise information among input feature channels using a cross-channel cross-covariance. This process generates an attention map that implicitly encodes the global context, facilitates interactions within the channels and promotes the color enhancement. The computational complexity of the channel-level global self-attention is linear, so the CRB can exhibit faster computational performance than spatial self-attention computations~\cite{ViT}. Fig.~\ref{fig:CRB} illustrates the specific structure of the CRB.

We used the Rescale Layer Normalization (RLN) from \cite{DehazeFormer} because it ensures consistency in the mean and variance between input and output feature maps. We believe that the RLN can help preserve the image brightness, restore colors, and reduce artifacts. For a given input $x\in {{\mathbb{R}}^{c\times h\times w}}$, we first applied the RLN and used the $1\times1$ point-wise convolution and $3\times3$ depth-wise convolution to map the RLN into the required $Q$, $K$, and $V$ values for self-attention computation. This process can be expressed by the following Eq.~\ref{equ:QKV}:
\begin{equation}\label{equ:QKV}
Q,K,V=Flat(DConv(PConv(RLN(x))))
\end{equation}

Here, $PConv(\cdot )$ is a $1\times1$ point-wise convolution; $DConv(\cdot)$ is a $3\times3$ depth-wise convolution; $Flat(\cdot)$ is a flattening operation; $Q,K,V\in {{\mathbb{R}}^{c\times hw}}$.

During the computation of the global channel-wise self-attention, we performed a $3\times3$ convolution operation on the $V$ values. This step helped improve the performance of the network by gathering information from the neighboring regions. The overall process of attention calculation can be described as follows:
\begin{equation}\label{equ:ChannelSelfAttention}
{Atten}=(softmax(Q\cdot {{K}^{T}})/\gamma )\cdot Conv3(V)
\end{equation}

\noindent where $\gamma$ is the scaling coefficient for stable computation, and $Conv3(\cdot)$ is the convolution operation with a $3\times3$ kernel size.

Finally, we restored the mean and variance (Affine) of $Atten$, added $Atten$ pixel-wise with the input $x$ and further refined the features through an MLP layer.

\subsubsection{Channel Fusion Block (CFB)}
\begin{figure}[htb]
    \centering
    \includegraphics[width=0.8\columnwidth]{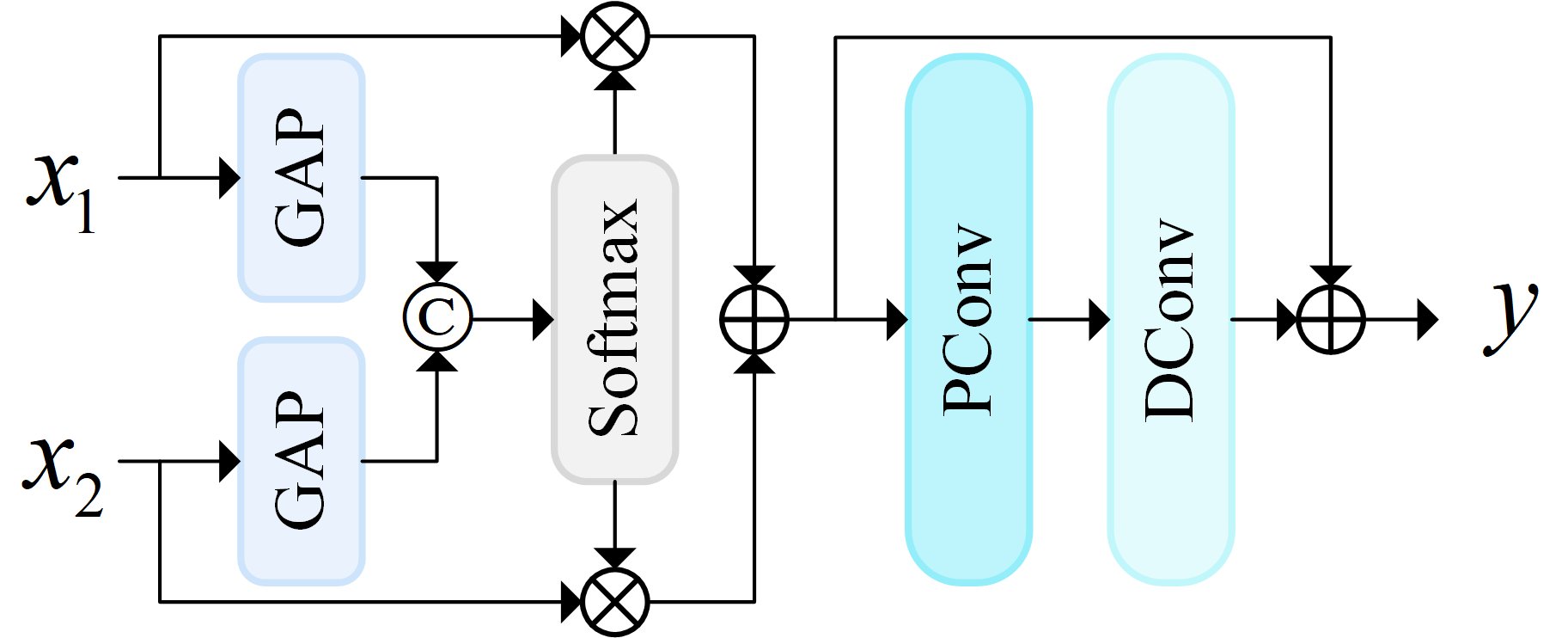}
    \caption{Overall structure of the CFB.}\label{fig:CFB}
\end{figure}

Pixel-wise addition and concatenation are the most common strategies for multi-feature map fusion. However, these methods treat information among different feature maps equally. We believe that due to the nonlinear absorption of light by water at different wavelengths, the information between channels in underwater image features is not equally important. Therefore, in our designed CFB, the information among different feature map channels is fused based on their importance.
Fig.~\ref{fig:CFB} shows the specific process of the CFB. For two given feature maps $x_{1}\in {{\mathbb{R}}^{c\times h\times w}}$ and $x_{2}\in {{\mathbb{R}}^{c\times h\times w}}$, we obtained the weights $\{{\alpha }_{1}$ and ${\alpha }_{2}\}$ using the global average pooling $GAP(\cdot)$, softmax function, and split operation. Then, we performed a weighted sum and refined the sum result using $PConv$, $DConv$, and residual connections. The more formal fusion process of the CFB is as follows:
\begin{equation}\label{equ:CFB}
\begin{aligned}
    & \{{{\alpha }_{1}},{{\alpha }_{2}}\}=split(softmax(GAP({{x}_{1}}),GAP({{x}_{2}}))) \\ 
    & {y}'={{\alpha }_{1}}\cdot {{x}_{1}}+{{\alpha }_{2}}\cdot {{x}_{2}} \\ 
    & y={y}'+DConv(PConv({y}')) \\ 
\end{aligned}
\end{equation}

\noindent where $y\in {{\mathbb{R}}^{c\times h\times w}}$ is the fused generated feature map.

\subsection{Loss Function}
The overall loss function of our WaterFormer contains three terms: the Reconstruction Loss ${{L}_{\ell_{1}}}$, Chromatic Consistency Loss ${{L}_{chroma}}$ and Sobel Color Loss ${{L}_{Sobel}}$.

\textbf{Reconstruction Loss.} For the given normalized real image ${{I}_{gt}}\in {{\mathbb{R}}^{h\times w\times c}}$ and network predicted image ${{I}_{pred}}\in {{\mathbb{R}}^{h\times w\times c}}$, ${{L}_{\ell_{1}}}$ is calculated as follows:
\begin{equation}\label{equ:L1Loss}
{{L}_{{{\ell }_{1}}}}=\frac{1}{h\times w\times c}\cdot \sum\limits_{k=1}^{c}{\sum\limits_{n=1}^{h}{\sum\limits_{m=1}^{w}{\left| {{I}_{gt}}(k,n,m)-{{I}_{pred}}(k,n,m) \right|}}}
\end{equation}

\textbf{Chromatic Consistency Loss.} To help the network learn the color differences between ${I}_{gt}$ and ${I}_{pred}$ while maintaining consistent chromaticity, we designed ${{L}_{chroma}}$ based on the chromatic similarity term proposed in \cite{FSIM}. First, we transformed ${I}_{gt}$ and ${I}_{pred}$ from the $RGB$ space to the $YIQ$ space using a matrix transformation:
\begin{equation}\label{equ:RGB2YIQ}
\left[ \begin{matrix}
   Y  \\
   I  \\
   Q  \\
\end{matrix} \right]=\left[ \begin{matrix}
   0.299 & 0.587 & 0.114  \\
   0.596 & -0.274 & -0.322  \\
   0.211 & -0.523 & 0.312  \\
\end{matrix} \right]\left[ \begin{matrix}
   R  \\
   G  \\
   B  \\
\end{matrix} \right]
\end{equation}

The $I$ and $Q$ chromaticity channels of image ${I}_{gt}$ (${I}_{pred}$) are represented by ${I}_{gt}^{I}$ (${I}_{pred}^{I}$) and ${I}_{gt}^{Q}$ (${I}_{pred}^{Q}$), respectively. The chromatic consistency values ${{S}_{chroma}^{I}}$ and ${{S}_{chroma}^{Q}}$ of channels $I$ and $Q$ are calculated as follows:
\begin{equation}\label{equ:ChromaConsistence}
\begin{aligned}
    &{{S}_{chroma}^{I}}=\frac{\sigma (I_{gt}^{I},I_{pred}^{I})+{{c}_{1}}}{{{\sigma }^{2}}(I_{gt}^{I})+{{\sigma }^{2}}(I_{pred}^{I})+{{c}_{1}}}\\
    &{{S}_{chroma}^{Q}}=\frac{\sigma (I_{gt}^{Q},I_{pred}^{Q})+{{c}_{2}}}{{{\sigma }^{2}}(I_{gt}^{Q})+{{\sigma }^{2}}(I_{pred}^{Q})+{{c}_{2}}}
\end{aligned}
\end{equation}

\noindent where $S_{chroma}^{I},S_{chroma}^{Q}\in \left( 0,1 \right]$, and the values closer to 1 indicate higher chromatic consistency between the two images. ${{\sigma }^{2}(\cdot)}$ is the variance calculation; ${{\sigma }(\cdot,\cdot)}$ is the covariance calculation; $c_{1}$ and $c_{2}$ are normalization constants to harmonize the chromatic consistency, which we set to 0.001 in our experiment. We defined $L_{chroma}$ as follows:
\begin{equation}\label{equ:ChromaLoss}
{L_{chroma}}=1-{S_{chroma}^{I}}\cdot{S_{chroma}^{Q}}
\end{equation}

In the experiment, to enhance the chromatic consistency of local regions, we set a sliding window with a size of $15\times15$ (see Tab.~\ref{tab:AblationStudyOfWindowSize}) and a stride of 1 to calculate the chromatic consistency loss for each local region. Finally, we obtained the total chromatic consistency loss by taking the mean of all local losses.

\textbf{Sobel Color Loss.} Hazing typically results in blurriness in underwater images, which manifests as the loss of edge details and a decrease in contrast. To overcome this challenge, we effectively preserved the fine color details between ${{I}_{gt}}$ and ${{I}_{pred}}$ by constraining the contour components of different color channels. Specifically, we defined the Sobel color loss in Eq.~\ref{equ:SobelLoss}:
\begin{equation}\label{equ:SobelLoss}
\begin{aligned}
    & L_{Sobel}^{x}=\frac{1}{3}\cdot \sum\limits_{\lambda }^{\{R,G,B\}}{{{L}_{{{\ell }_{1}}}}(Sobe{{l}_{x}}(I_{gt}^{\lambda }),Sobe{{l}_{x}}(I_{pred}^{\lambda }))} \\ 
    & L_{Sobel}^{y}=\frac{1}{3}\cdot \sum\limits_{\lambda }^{\{R,G,B\}}{{{L}_{{{\ell }_{1}}}}(Sobe{{l}_{y}}(I_{gt}^{\lambda }),Sobe{{l}_{y}}(I_{pred}^{\lambda }))} \\ 
    & {{L}_{Sobel}}=L_{Sobel}^{x}+L_{Sobel}^{y}
\end{aligned}
\end{equation}

\noindent where $Sobe{{l}_{x}}(\cdot)$ and $Sobe{{l}_{y}}(\cdot)$ are the convolution operations of $Sobel$ operator in the $x$ and $y$ directions, respectively. The final loss function is defined as:
\begin{equation}\label{equ:LossFunction}
L={{\lambda }_{1}}\cdot {{L}_{{{\ell }_{1}}}}+{{\lambda }_{2}}\cdot {{L}_{chroma}}+{{\lambda }_{3}}\cdot {{L}_{Sobel}}
\end{equation}

Here, ${\lambda }_{1}$, ${\lambda }_{2}$, and ${\lambda }_{3}$ are constants that balance the proportion of each loss term, set to 1, 1, and 2, respectively, based on numerous experiments (see Tab.~\ref{tab:AblationStudyOfLossTerm}).

\section{Experiments}
\label{sec:experiments}
In this section, we present the details and settings of the experiments. Then, we evaluate the performance of the proposed model and compare it with other representative methods. Finally, we conduct a series of ablation experiments to demonstrate the effectiveness of the proposed components.

\subsection{Implementation Details}
Our experiments were conducted on both synthetic underwater datasets~\cite{UWCNN} and real underwater dataset UIEBD~\cite{WaterNet_UIEBD}. In the synthetic underwater dataset, the reference images followed real color distributions, whereas the synthetic images simulated 10 different underwater degradation environments (I, IA, IB, II, and III simulated open sea underwater environments; 1, 3, 5, 7, and 9 simulated nearshore underwater environments). Each category of synthetic images had 1449 images, which were paired with reference images. We randomly selected 300 paired images from each category as the training set.
The UIEBD dataset includes 890 paired underwater real images and reference images, as well as 60 unpaired underwater real images. The reference images were obtained through enhancement using various traditional methods. We randomly partitioned 800 paired images as the training set.
All images were resized to $256\times256$ for training, and we augmented the dataset by applying random flips and rotations.

The experiments were conducted on tow NVIDIA GPUs (GeForce GTX 2080Ti) using the PyTorch 1.7.1 framework on an Ubuntu 16.04 system. 
We adopt the AdamW optimizer with hyperparameters ${\beta}_{1}=0.9$ and ${\beta}_{2}=0.999$, initializing the learning rate at $1\times 10^{-3}$. A cosine annealing scheduler (CosineAnnealingLR) progressively adjusts the learning rate from $1\times 10^{-3}$ to $1\times 10^{-7}$ over 400 training epochs, with a batch size of 4.

\subsection{Experiment settings} 
\textbf{Benchmarks.} In the synthetic underwater dataset, we randomly selected 30 additional paired images from each synthetic data category as the test set, which was denoted Test-S300. In the UIEBD dataset, we selected the remaining 90 paired images as the test set, which was denoted as Test-U90. Test-S300 and Test-U90 served as our full-reference benchmarks for the experiments.
To demonstrate the generalization performance of our proposed network, we randomly selected 60 and 16 unpaired real underwater images from the UIEBD dataset and SQUID dataset~\cite{SQUID}, respectively, as the no-reference datasets. These datasets are called Test-U60 and Test-SQ16, respectively.

\textbf{Compared Methods.} We compared the proposed WaterFormer to other state-of-the-art UIE methods based on deep learning: \UWCNN, \WaterNet, \Ucolor, \FUnIE, \Ushape, \UDAformer, \GuidedHybSensUIR, \DMwater, and \WFDiff. In addition, we compared WaterFormer with the following traditional methods: \Fusion, \Retinex, \MLLE, \UDCP, \RED, and \HFM.

\textbf{Evaluation Metrics.}
For Test-S300 and Test-U90, we used SSIM, PSNR, and NRMSE as the full-reference evaluation metrics for our experiments. These metrics are widely used to assess the similarity between two images due to their convenience and accuracy. Higher values of SSIM and PSNR or lower values of NRMSE indicate greater similarity between two images.
For Test-U60, we used UCIQEE~\cite{UCIQE} and UIQM~\cite{UIQM} as no-reference evaluation metrics. Higher values of UCIQE and UIQM indicate better visual perception of the underwater images by humans. However, UCIQE and UIQM may not accurately reflect actual visual perception in certain cases.

\begin{table*}[htb]
\centering
\scalebox{0.9}{
\begin{tabular}{c||ccc||ccc||ccc}
\hline
\multirow{2}{*}{\textbf{Methods}} & \multicolumn{3}{c||}{\textbf{Test-S300}}         & \multicolumn{3}{c||}{\textbf{Test-U90}}          & \multicolumn{3}{c}{Overhead}                     \\ \cline{2-10} 
                                  & SSIM↑          & PSNR↑          & NRMSE↓         & SSIM↑          & PSNR↑          & NRMSE↓         & \#Param(M)↓    & MACs(G)↓       & Time(s)↓       \\ \hline
\Fusion                           & 0.651          & 12.65          & 0.685          & 0.872          & 21.80          & 0.188          & ×              & ×              & 0.073          \\
\Retinex                          & 0.641          & 14.54          & 0.520          & 0.773          & 17.34          & 0.283          & ×              & ×              & 0.208          \\
\MLLE                             & 0.603          & 12.38          & 0.667          & 0.740          & 16.50          & 0.315          & ×              & ×              & 0.038          \\
\UDCP                             & 0.597          & 13.84          & 0.500          & 0.689          & 13.85          & 0.418          & ×              & ×              & 1.694          \\
\RED                              & 0.659          & 12.66          & 0.674          & 0.833          & 19.03          & 0.244          & ×              & ×              & 0.020          \\
\HFM                              & 0.676          & 15.00          & 0.377          & 0.835          & 18.32          & 0.293          & ×              & ×              & 0.328          \\
\UWCNN                            & 0.640          & 14.62          & 0.456          & 0.818          & 18.10          & 0.265          & \textbf{0.041} & \textbf{2.649} & \textbf{0.002} \\
\WaterNet                         & 0.746          & 19.25          & 0.322          & 0.905          & 23.17          & 0.153          & 1.091          & 71.42          & 0.016          \\
\FUnIE                            & 0.769          & 21.34          & 0.231          & 0.814          & 18.94          & 0.239          & 7.020          & 10.24          & \textbf{0.002} \\
\Ucolor                           & 0.813          & 22.23          & 0.235          & 0.897          & 21.23          & 0.180          & 148.8          & 404.8          & 0.169          \\
\Ushape                           & 0.793          & 21.89          & 0.213          & 0.806          & 20.28          & 0.208          & 22.82          & {\ul 2.983}    & 0.025          \\
\UDAformer                        & 0.798          & 22.39          & 0.226          & 0.921          & 23.38          & 0.147          & 9.594          & 41.59          & 0.042          \\
\DMwater                          & 0.867          & 25.29          & 0.169          & {\ul 0.926}    & \textbf{25.32} & \textbf{0.131} & 10.70          & 66.89          & 0.166          \\
\WFDiff                           & 0.827          & 22.41          & 0.234          & 0.924          & 23.82          & 0.144          & 18.48          & 122.76         & 0.465          \\
\GuidedHybSensUIR                 & {\ul 0.902}    & {\ul 27.52}    & {\ul 0.131}    & {\ul 0.926}    & 24.19          & 0.136          & 1.145          & 10.05          & 0.104          \\
WaterFormer                       & \textbf{0.917} & \textbf{31.18} & \textbf{0.105} & \textbf{0.928} & {\ul 24.65}    & {\ul 0.132}    & {\ul 0.313}    & 7.749          & 0.023          \\ \hline
\end{tabular}
}
\caption{Quantitative comparison results for each method on the Test-S300 and Test-U90 datasets. Bold and underline indicate the $1^{st}$ and $2^{nd}$ ranks, respectively.} \label{tab:CompareMethods}
\end{table*}

\subsection{Results and Analysis}
\textbf{Full-reference Evaluation.} Tab.~\ref{tab:CompareMethods} shows the quantitative comparison results on Test-S300 and Test-U90. The optimal results are presented in bold, while the second-best results are marked with an underline. WaterFormer achieved excellent SSIM, PSNR, and NRMSE scores on both full-reference test sets.
What is noteworthy is that, for the Test-S300 dataset, WaterFormer exceeded the second-best method by 1.66\% in SSIM, 13.3\% in PSNR, and achieved a 19.8\% lower NRMSE. This significant improvement indicated that previous methods were not able to effectively restore the true color distribution, detail distribution, and noise distribution of underwater images. 
For the Test-U90 dataset, WaterFormer achieved the best performance in SSIM, surpassing the second-best method, DM-water, by 0.22\%. However, in terms of PSNR and NRMSE, WaterFormer ranked second, being 2.65\% lower and 0.76\% higher than DM-water, respectively. SSIM considers the consistency of brightness, contrast, and structural information, indicating that WaterFormer focused more on results aligned with human visual perception. In contrast, PSNR and NRMSE are more sensitive to pixel-level noise, suggesting a greater emphasis on noise reduction capability. Despite slight differences in their inferred results, the evaluation metrics were very close, demonstrating WaterFormer's exceptional ability in image enhancement and fitting.

\begin{figure*}[p]
\centering
   \begin{overpic}[width=\textwidth]{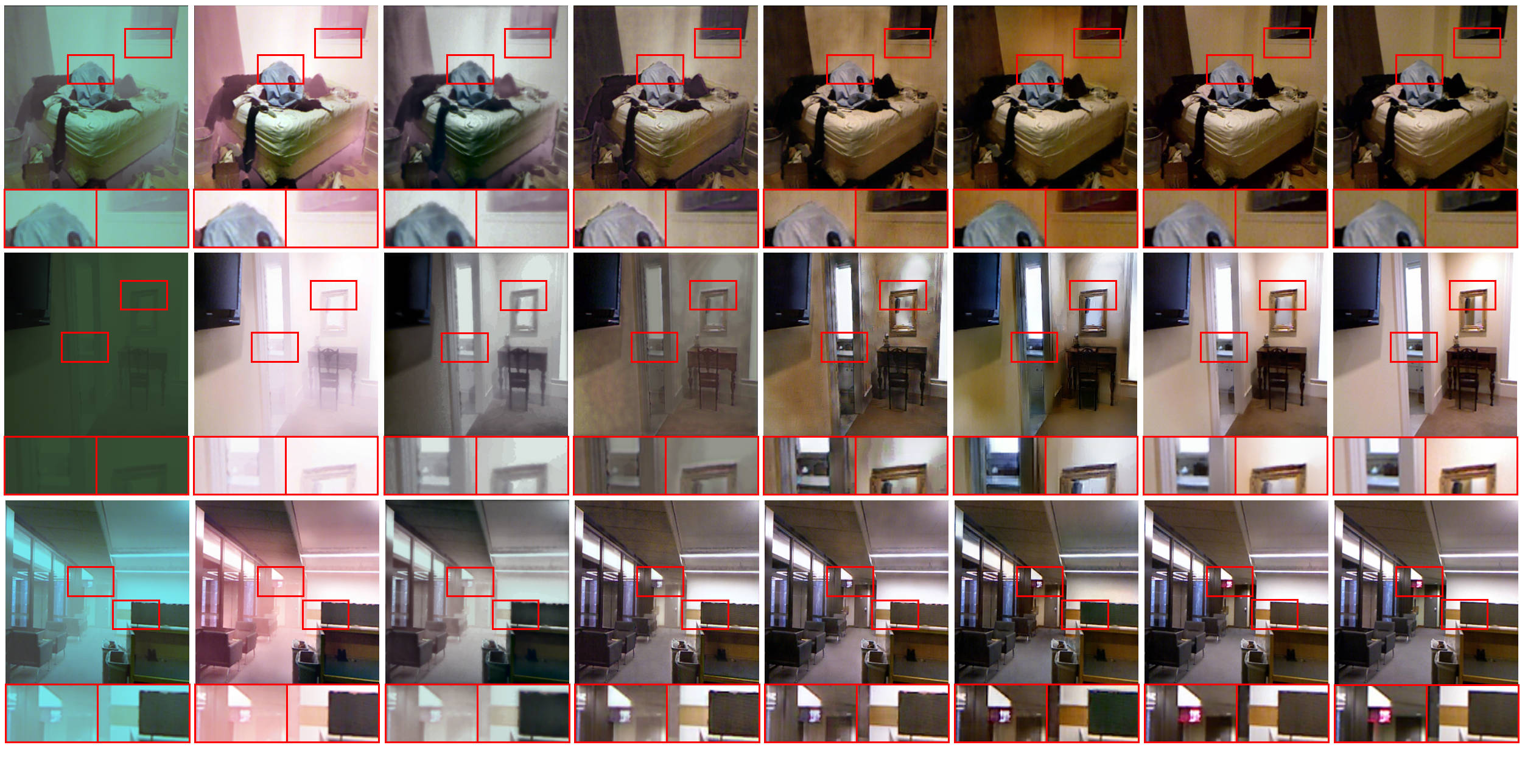} \small
    \put(5,0){(a)}
    \put(17.5,0){(b)}
    \put(30,0){(c)}
    \put(42.3,0){(d)}
    \put(55,0){(e)}
    \put(67.5,0){(f)}
    \put(79.5,0){(g)}
    \put(92,0){(h)}
   \end{overpic}
   \caption{Qualitative Comparison Results on Test-S300. (a) Input, (b)~\Fusion, (c)~\Retinex, (d)~\WaterNet, (e)~\Ucolor, (f)~\WFDiff, (g) WaterFormer, (h) GT. 
   }\label{fig:VisualComparison1}
\end{figure*}

\begin{figure*}[p]
\centering
   \begin{overpic}[width=\textwidth]{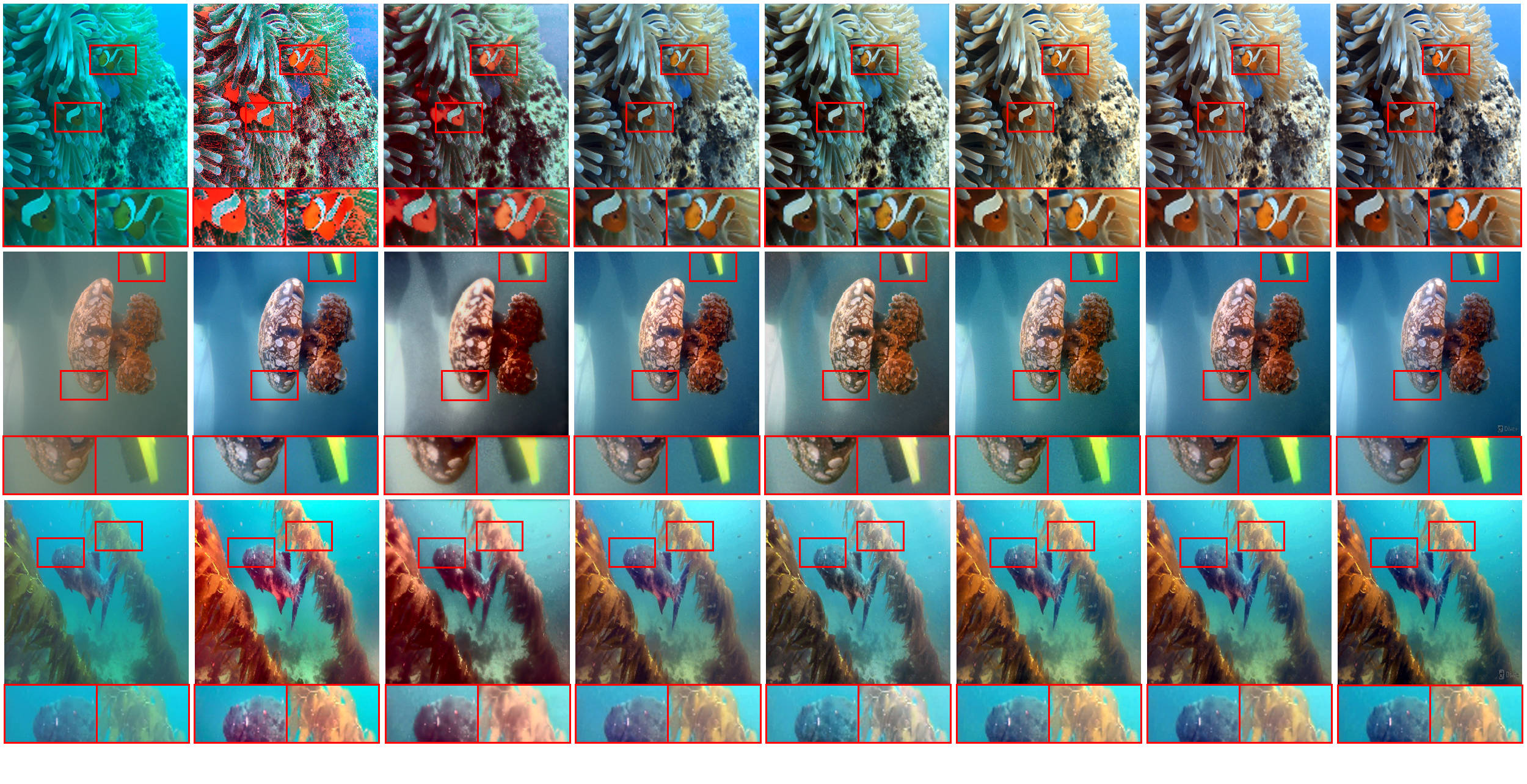} \small
    \put(5,0){(a)}
    \put(17.5,0){(b)}
    \put(30,0){(c)}
    \put(42.3,0){(d)}
    \put(55,0){(e)}
    \put(67.5,0){(f)}
    \put(79.5,0){(g)}
    \put(92,0){(h)}
   \end{overpic}
   \caption{Qualitative Comparison Results on Test-U90. (a) Input, (b)~\Fusion, (c)~\Retinex, (d)~\WaterNet, (e)~\Ucolor, (f)~\WFDiff, (g) WaterFormer, (h) GT. 
   }\label{fig:VisualComparison2}
\end{figure*}

Fig.~\ref{fig:VisualComparison1} and Fig.~\ref{fig:VisualComparison2} show the qualitative comparison results on Test-S300 and Test-U90, respectively. 
For the Test-S300 dataset, images enhanced using Fusion-based and Retinex-based methods exhibited noticeable reddish and blackish color biases, failing to recover the true details obscured by haze. Although the Water-Net, Ucolor, and WF-Diff methods succeeded in roughly restoring the original colors of the synthesized images, they still fell short of recovering the full saturation and tonal distribution. Furthermore, they introduced color artifacts that were absent in the original images. In contrast, WaterFormer did not produce such artifacts and successfully restored brightness, color tones, and details across different regions of the enhanced images.
Regarding the Test-U90 dataset, Fusion-based and Retinex-based enhancements still resulted in red and black color biases, along with prominent red edge artifacts. The ability of Ucolor to recover colors was relatively limited, and the enhanced images lacked sufficient tone and saturation. However, the results from Water-Net, WF-Diff, and WaterFormer were closer to the reference images, demonstrating their effectiveness in accurately restoring both color and fine details from real underwater scenes.

\textbf{Non-reference Evaluation.} Tab.~\ref{tab:CompareMethods2} shows the quantitative comparison results on Test-U60 and Test-SQ16. The best and second best results are presented in bold and underlined, respectively. Unfortunately, our proposed WaterFormer did not achieve the best or second-best results in the UIQM and UCIQE metrics on either no-reference datasets.
As noted in ~\cite{Ucolor,u-shape}, while the UIQM and UCIQE metrics effectively evaluate the color, brightness, contrast, saturation, and sharpness of underwater images, they struggle to accurately measure color artifacts and shifts. In scenarios with more color artifacts and shifts, these metrics tend to yield higher scores, contradicting human perception. Our WaterFormer consistently avoids color artifacts and shifts, contributing to our lower UIQM and UCIQE scores. Hence, while we consider the UIQM and UCIQE scores as indicative, they should not be the sole criteria for assessing the algorithm performance.

\begin{table}[htb]
\centering
\scalebox{0.80}{
\begin{tabular}{c||cc||cc}
\hline
\multirow{2}{*}{\textbf{Methods}} & \multicolumn{2}{c||}{\textbf{Test-U60}} & \multicolumn{2}{c}{\textbf{Test-SQ16}} \\ \cline{2-5} 
                                  & UIQM↑              & UCIQE↑            & UIQM↑              & UCIQE↑            \\ \hline
Input                             & 2.042              & 0.478             & 0.813              & 0.393             \\
\Fusion                           & 2.601              & \textbf{0.621}    & 1.629              & 0.589             \\
\Retinex                          & 2.677              & 0.582             & 2.336              & 0.597             \\
\MLLE                             & 1.967              & 0.593             & 2.340              & 0.558             \\
\UDCP                             & 1.639              & 0.554             & 1.067              & {\ul 0.599}       \\
\RED                              & 2.330              & 0.559             & 1.362              & 0.560             \\
\HFM                              & 2.298              & 0.585             & 2.273              & \textbf{0.607}    \\
\UWCNN                            & 2.411              & 0.487             & 2.018              & 0.404             \\
\WaterNet                         & 2.694              & {\ul 0.598}       & {\ul 2.562}        & 0.567             \\
\FUnIE                            & \textbf{2.927}     & 0.560             & \textbf{2.599}     & 0.491             \\
\Ucolor                           & 2.695              & 0.564             & 2.383              & 0.508             \\
\Ushape                           & {\ul 2.812}        & 0.528             & 2.184              & 0.493             \\
\UDAformer                        & 2.807              & 0.576             & 2.487              & 0.519             \\
\DMwater                          & 2.704              & 0.575             & 2.212              & 0.525             \\
\WFDiff                           & 2.785              & 0.576             & 2.336              & 0.520             \\
\GuidedHybSensUIR                 & 2.757              & 0.584             & 2.440              & 0.556             \\
WaterFormer                       & 2.594              & 0.581             & 2.395              & 0.551             \\ \hline
\end{tabular}
}
\caption{Quantitative comparison results for each method on the Test-U60 and Test-SQ16 datasets. Bold and underline indicate the $1^{st}$ and $2^{nd}$ ranks, respectively.} \label{tab:CompareMethods2}
\end{table}

\begin{figure*}[p]
\centering
   \begin{overpic}[width=\textwidth]{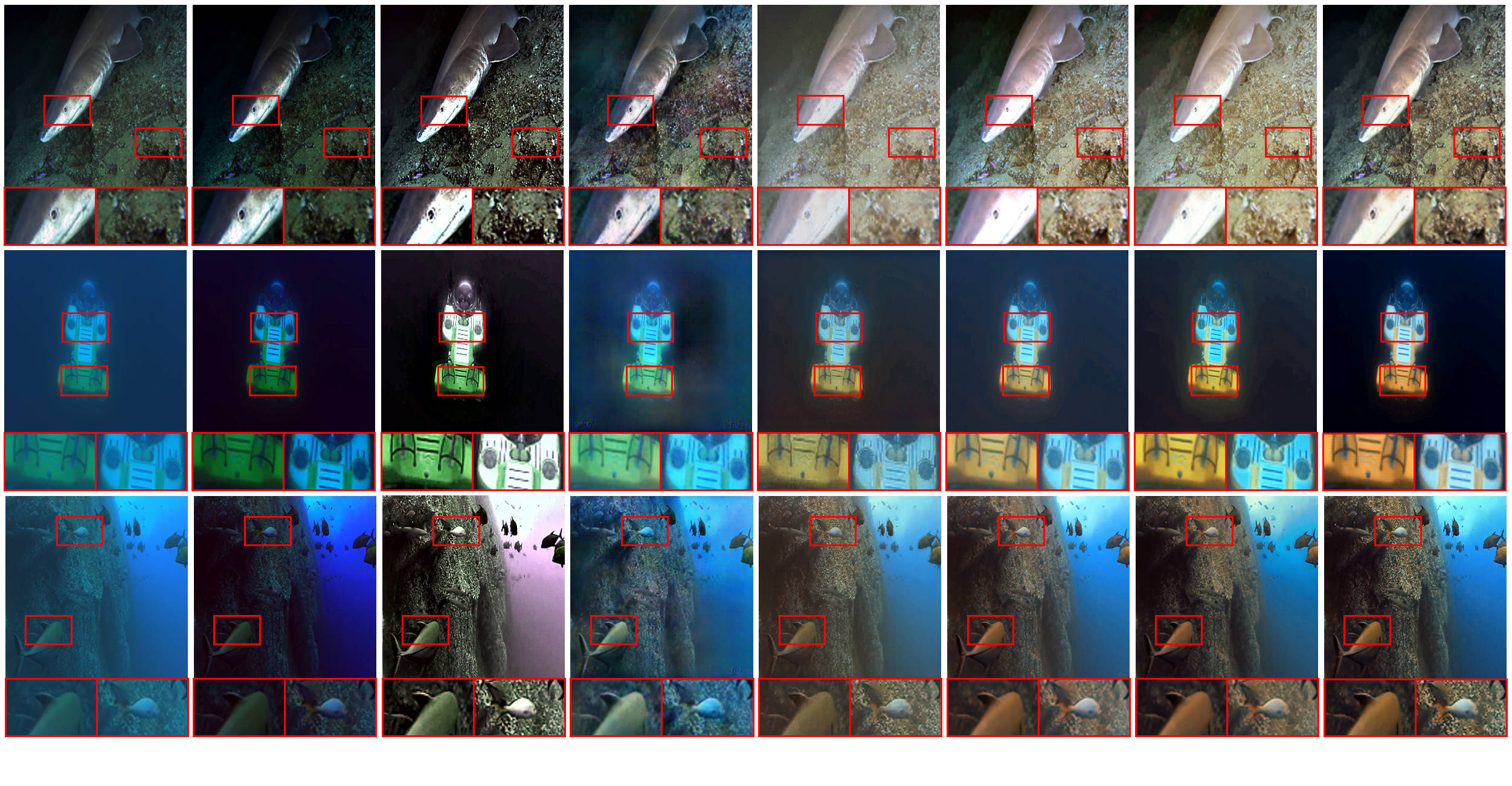} \small
    \put(5,0){(a)}
    \put(17.5,0){(b)}
    \put(30,0){(c)}
    \put(42.3,0){(d)}
    \put(55,0){(e)}
    \put(67.5,0){(f)}
    \put(79.5,0){(g)}
    \put(92,0){(h)}
   \end{overpic}
   \caption{Qualitative Comparison Results on Test-U60. (a) Input, (b)~\UDCP, (c)~\MLLE, (d)~\FUnIE, (e)~\Ushape, (f)~\DMwater, (g)~\GuidedHybSensUIR, (h) WaterFormer.
   }\label{fig:VisualComparison3}
\end{figure*}

\begin{figure*}[p]
\centering
   \begin{overpic}[width=\textwidth]{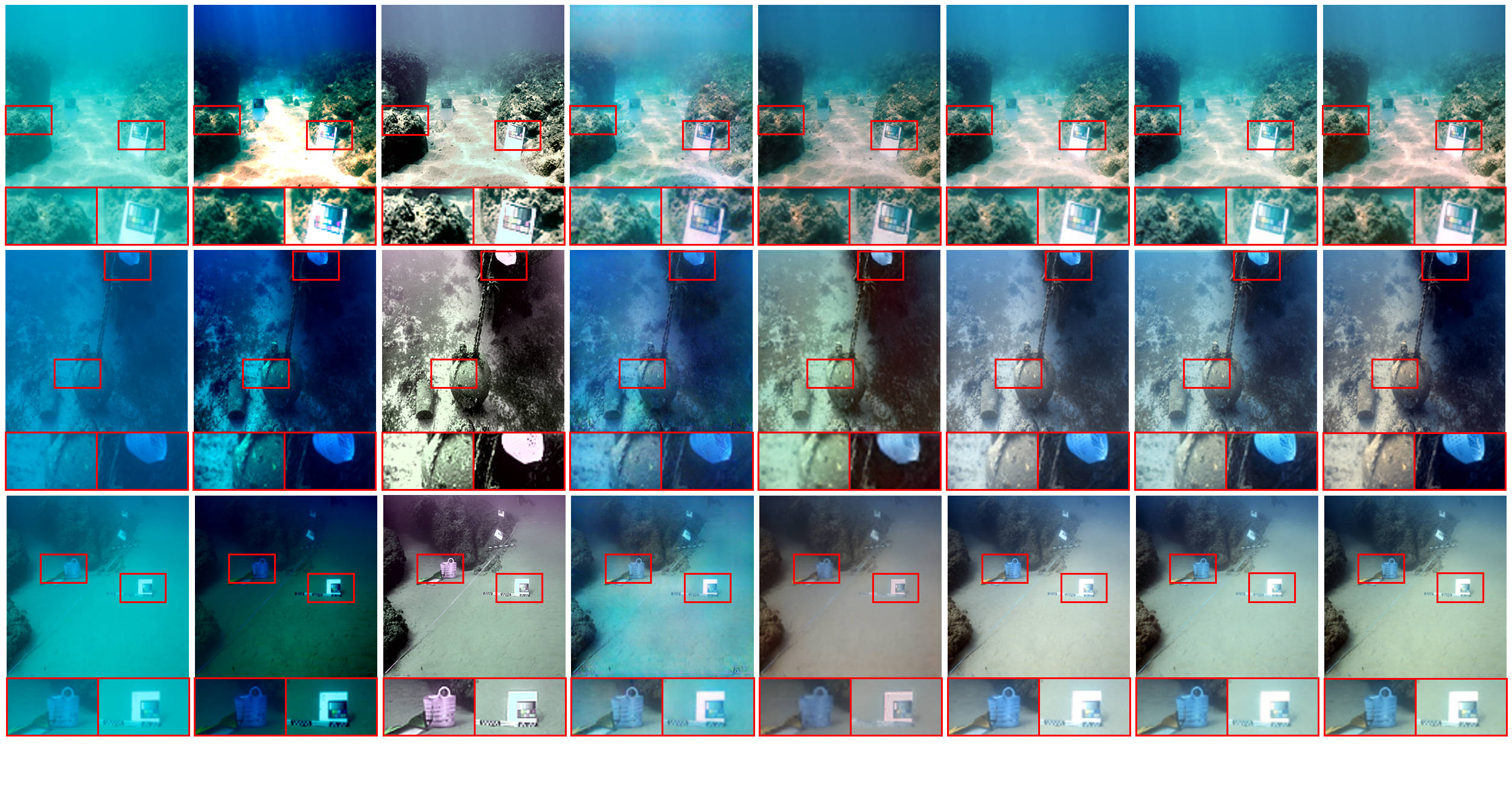} \small
    \put(5,0){(a)}
    \put(17.5,0){(b)}
    \put(30,0){(c)}
    \put(42.3,0){(d)}
    \put(55,0){(e)}
    \put(67.5,0){(f)}
    \put(79.5,0){(g)}
    \put(92,0){(h)}
   \end{overpic}
   \caption{Qualitative Comparison Results on Test-SQ16. (a) Input, (b)~\UDCP, (c)~\MLLE, (d)~\FUnIE, (e)~\Ushape, (f)~\DMwater, (g)~\GuidedHybSensUIR, (h) WaterFormer.
   }\label{fig:VisualComparison4}
\end{figure*}

\begin{figure*}[p]
\centering
   \begin{overpic}[width=\textwidth]{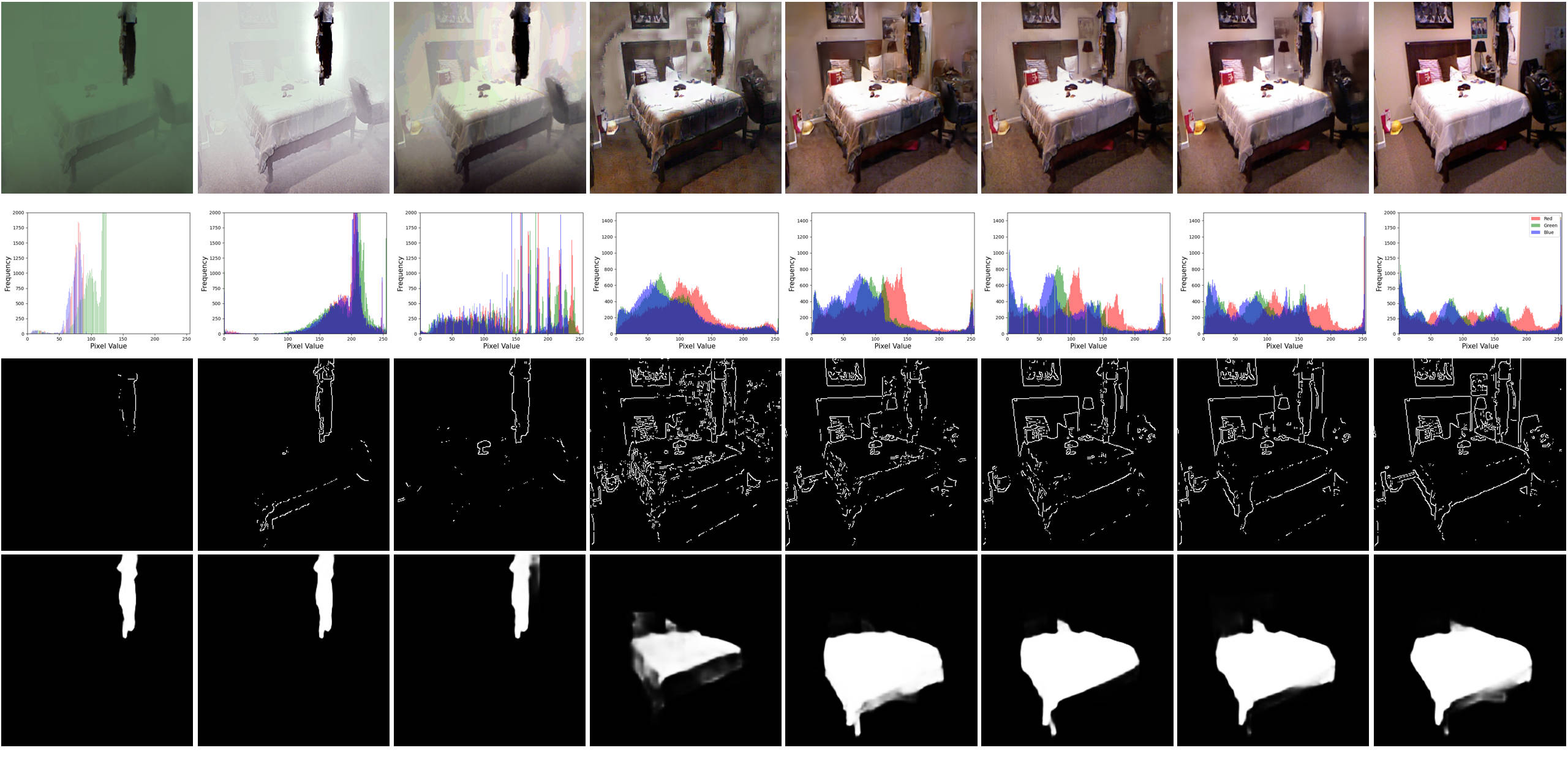} \small
    \put(4.5,-0.5){(a)}
    \put(17,-0.5){(b)}
    \put(29.5,-0.5){(c)}
    \put(42,-0.5){(d)}
    \put(54.5,-0.5){(e)}
    \put(67,-0.5){(f)}
    \put(79.5,-0.5){(g)}
    \put(92,-0.5){(h)}
   \end{overpic}
   \caption{Rows 1-4 show the enhanced results, histogram display results, Canny edge detection results, and saliency object detection results on Test-S300, respectively. (a) Input, (b)~\MLLE, (c)~\HFM, (d)~\UDAformer, (e)~\DMwater, (f)~\GuidedHybSensUIR, (g)WaterFormer, (h) GT.
   }\label{fig:VisualComparison6}
\end{figure*}

\begin{figure*}[p]
\centering
   \begin{overpic}[width=\textwidth]{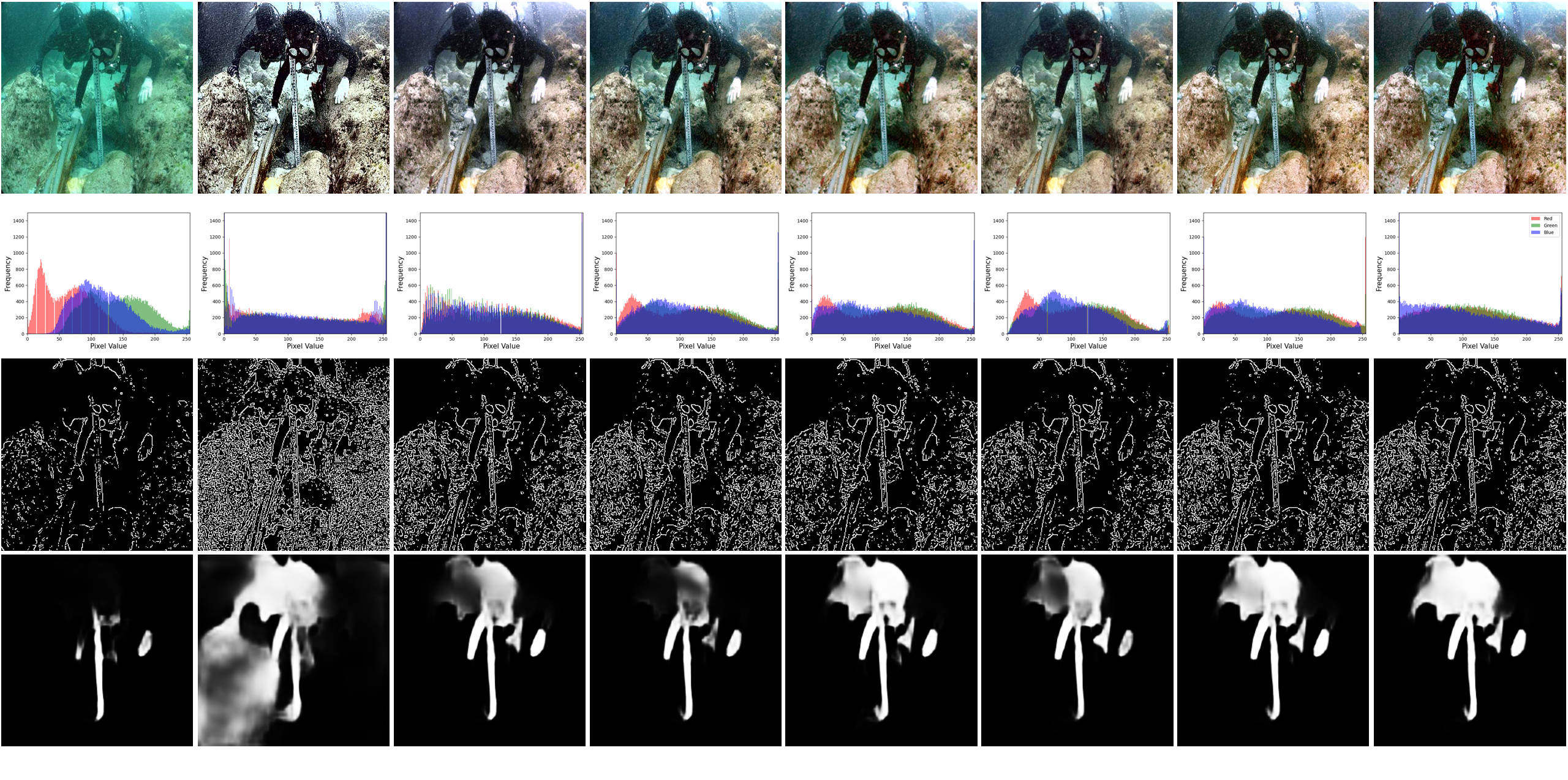} \small
    \put(4.5,-0.5){(a)}
    \put(17,-0.5){(b)}
    \put(29.5,-0.5){(c)}
    \put(42,-0.5){(d)}
    \put(54.5,-0.5){(e)}
    \put(67,-0.5){(f)}
    \put(79.5,-0.5){(g)}
    \put(92,-0.5){(h)}
   \end{overpic}
   \caption{Rows 1-4 show the enhanced results, histogram display results, Canny edge detection results, and saliency object detection results on Test-U90, respectively. (a) Input, (b)~\MLLE, (c)~\HFM, (d)~\UDAformer, (e)~\DMwater, (f)~\GuidedHybSensUIR, (g)WaterFormer, (h) GT.
   }\label{fig:VisualComparison7}
\end{figure*}

Fig.~\ref{fig:VisualComparison3} and Fig.~\ref{fig:VisualComparison4} show the qualitative comparison results on Test-U60 and Test-SQ16, respectively. 
For the Test-U60 dataset, none of the methods produced satisfactory results. For example, while the UDCP method performed color correction, its capability was relatively weak, and the resulting images suffered from insufficient brightness. The MLLE method, on the other hand, produced incorrect color recovery results, such as a green submarine body (second row) and a purple ocean (third row). The FUnIE method introduced color artifacts that were inaccurate. All of the U-shape, DM-water, and GuidedHybSensUIR methods produced enhanced images that lacked adequate contrast and sharpness. Unlike the above methods, the images restored by WaterFormer exhibited satisfactory brightness, contrast, sharpness, and color.
Similarly, for the Test-SQ16 dataset, the UDCP method again produced images with insufficient brightness. The MLLE and FUnIE methods struggled with recovering the correct colors and introduced severe color artifacts. The U-shape, DM-water and GuidedHybSensUIR methods also failed to restore the contrast and sharpness of the images adequately. In comparison, the images generated by WaterFormer restored more colors and details as much as possible, without introducing any additional color shifts or artifacts.

\textbf{Overhead.} In Tab.~\ref{tab:CompareMethods}, we included the inference speed of each method. Furthermore, for deep learning methods, we measured the total cost of the network using the total number of parameters (\#Param) and multiply-accumulate operations (MACs). All inference speed and network cost measurements were conducted on images of size $256\times256$. Compared to other methods, our proposed WaterFormer achieves optimal inference results while maintaining fast inference speed and relatively low network overhead.

\subsection{Other Comparisons}
To further demonstrate that our proposed method has better color correction and broader applications than other methods, qualitative analyses were conducted on Test-S300 and Test-U90. Each image underwent histogram display, Canny edge detection, and saliency object detection. The saliency object detection method was referenced from~\cite{F3Net}, and Fig.~\ref{fig:VisualComparison6} and Fig.~\ref{fig:VisualComparison7} show the corresponding results. For Test-S300 and Test-U90, MLLE, HFM, and UDAformer failed to accurately reproduce the color, gradient, and saliency features of the reference images. For example, UDAformer introduced color artifacts in Test-S300, leading to inaccurate edge detection results. MLLE applied uniform processing across all channels in Test-U90, introducing gradient noise and reducing saliency features. Conversely, DM-water, GuidedHybSensUIR, and WaterFormer restored the information more accurately. Upon closer inspection, WaterFormer provided the most precise results. For instance, WaterFormer showed a more accurate color distribution in Test-S300, and it yielded results similar to DM-water in Test-U90, while being closer to the reference image in terms of color histogram and saliency feature detection results. In summary, WaterFormer exhibited outstanding performance in the UIE task and proves to be the most beneficial in related applications.

\subsection{Ablation Study}
The proposed WaterFormer incorporates four main components: the network backbone, CRB, CFB, and the loss function formed by ${{L}_{\ell_{1}}}$, ${{L}_{chroma}}$ and ${{L}_{Sobel}}$. To validate the effectiveness of each component, a series of ablation studies were conducted on the baseline framework with different combinations of these components. The Test-S300 was used for the ablation experiments, with each image resized to $256\times256$. The experiments were divided into four parts: first, WaterFormer was broken down into various variants, each composed of different components, and tested on Test-S300; second, the superiority of the CRB design was further explored, and different attention modules were tested as replacements for CRB; third, the CFB fusion mechanism was visualized, with pixel-wise addition used as a comparison; fourth, other optimizations and experimental parameters of WaterFormer were tested by ablation.

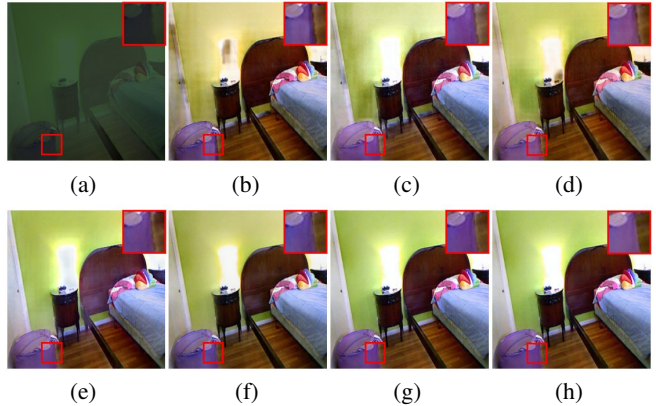
\begin{figure}[htb]
    \centering
   \begin{overpic}[width=\columnwidth]{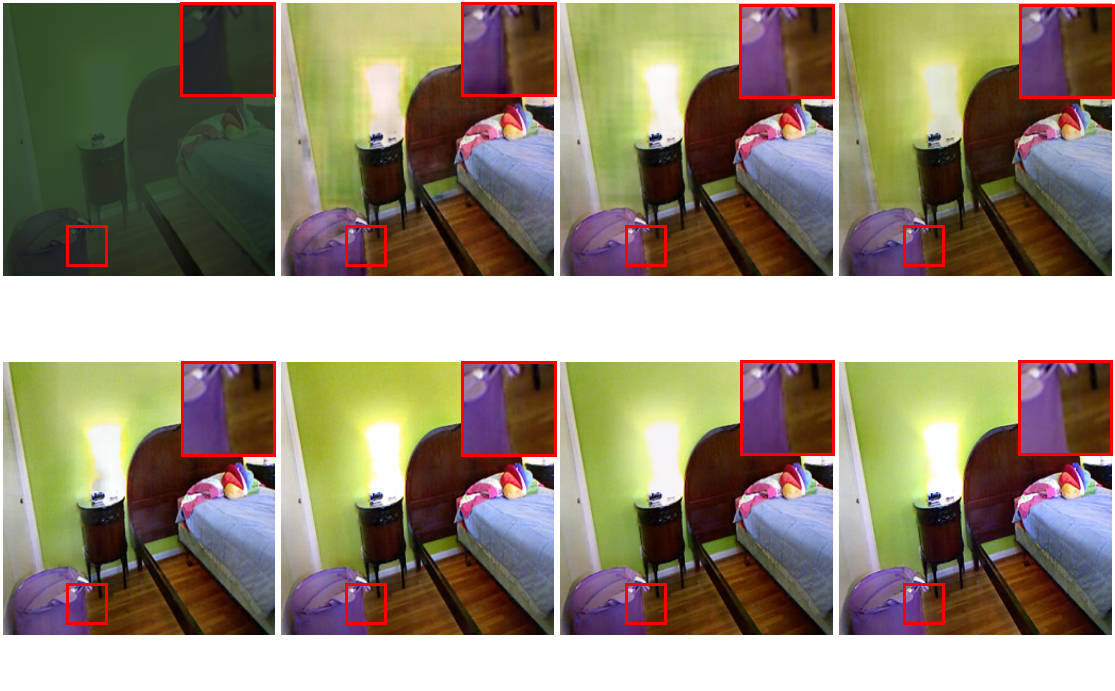} \small
    \put(10,32){(a)}
    \put(35,32){(b)}
    \put(60,32){(c)}
    \put(85,32){(d)}
    \put(10,0){(e)}
    \put(35.5,0){(f)}
    \put(60,0){(g)}
    \put(85,0){(h)}
   \end{overpic}
   \caption{Qualitative analysis of the WaterFormer ablation experiments. (a) Input, (b) Base, (c) $V_{1}$, (d) $V_{2}$, (e) $V_{3}$, (f) $V_{4}$, (g) $V_{5}$, (h) GT.
   }\label{fig:AblationStudy}
\end{figure}

\begin{table}[htb]
\centering
\scalebox{0.85}{
\begin{tabular}{c||cccc||cc}
\hline
Variants    & CRB & CFB & ${{L}_{chroma}}$ & ${{L}_{Sobel}}$     & SSIM↑   & PSNR↑ \\ \hline
Base        & w/o & w/o & w/o              & w/o                 & 0.876   & 27.32 \\
${{V}_{1}}$ & \checkmark & w/o & w/o       & w/o                 & 0.890   & 28.50 \\
${{V}_{2}}$ & \checkmark   & \checkmark    & w/o      & w/o      & 0.892   & 28.87 \\
${{V}_{3}}$ & \checkmark   & \checkmark    & \checkmark   & w/o  & 0.907   & 30.35 \\
${{V}_{4}}$ & \checkmark   & \checkmark    & w/o   & \checkmark  & 0.913   & 30.46 \\
${{V}_{5}}$ & \checkmark   & \checkmark    & \checkmark    & \checkmark    & \textbf{ 0.917}  & \textbf{ 31.18} \\ \hline
\end{tabular}
}
\caption{Quantitative analysis of the WaterFormer ablation experiments. The bold font indicates the best performance.} \label{tab:AblationStudy}
\end{table}

\textbf{Variants Evaluation.} Fig.~\ref{fig:AblationStudy} and Tab.~\ref{tab:AblationStudy} display the statistical results and visualization results on the Test-S300. The base model was composed only of the backbone and was trained using the ${{L}_{\ell_{1}}}$. It had basic UIE capabilities but still retained significant shortcomings in addressing color degradation and edge artifacts. 
The $V_{1}$ model introduced the CRB, which enabled it to recover color more accurately. However, due to insufficient integration of different features, the results of $V_{1}$ still exhibited a large amount of color artifacts.
Therefore, by introducing the CFB, the predictions of the $V_{2}$ model exhibited fewer color artifacts and a more realistic color distribution. The addition of ${{L}_{chroma}}$ to the $V_{3}$ model further improved the consistency between inference results and the chromaticity of real images. However, $V_{1}$, $V_{2}$, and $V_{3}$ continued to exhibit noticeable edge artifacts, such as the white rim of the bucket, and this problem was addressed by including ${{L}_{Sobel}}$, as demonstrated in the results of the inference $V_{4}$. Finally, by combining all components, our WaterFormer ($V_{5}$) yielded the closest results to the real images with the highest SSIM and PSNR scores.

\begin{figure}[htb]
    \centering
   \begin{overpic}[width=\columnwidth]{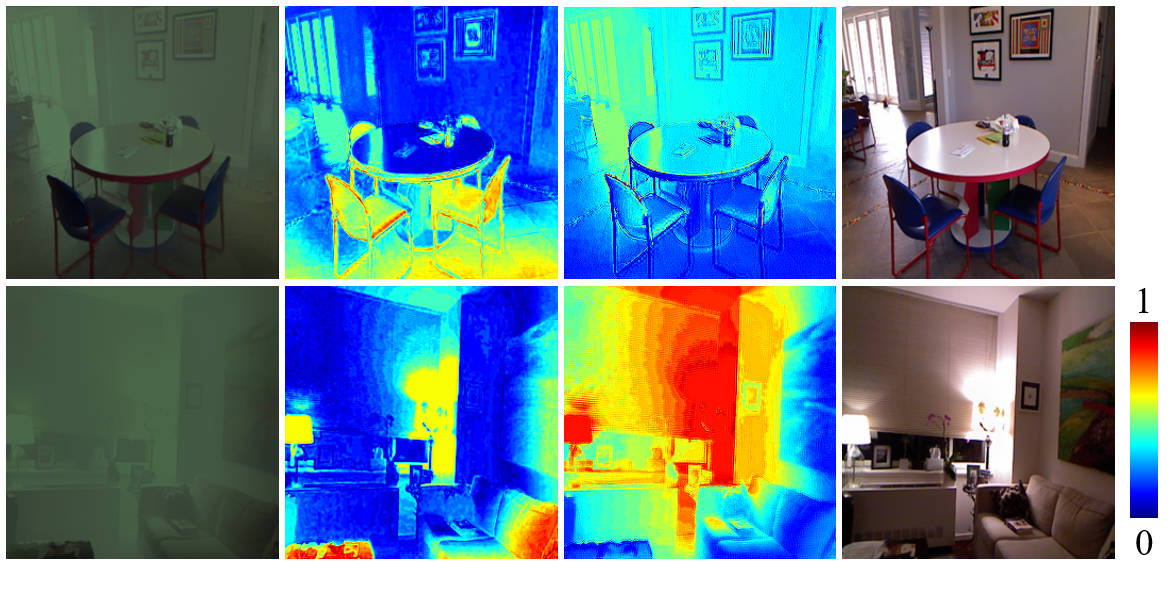} \small
    \put(9,0){(a)}
    \put(33,0){(b)}
    \put(57,0){(c)}
    \put(80,0){(d)}
   \end{overpic}
   \caption{Visualization of the CRB working mechanism. (a) Input, (b) result of DehazeFormer Block, (c) result of CRB, (d) GT.
   }\label{fig:AblationStudyOfCRB}
\end{figure}

\textbf{CRB Evaluation.} The design of CRB is based on the mechanism of self-attention between channels, as proposed by ~\cite{Restormer} and ~\cite{DNF}. CRB extends this mechanism and applies it effectively to the UIE task.
To visualize the effect of CRB, ~\ref{fig:AblationStudyOfCRB} shows the output of the feature map of the first 2-block CRB in the $V_{1}$ variant. For comparison, the feature map output of the first 2-block DehazeFormer Block in the Base model is included. We used maximum projection and the jet color map to plot the heatmap, where blue represents lower attention from the network, and red represents higher attention. In Test-S300, we selected two representative types of water for validation. The first row represents clear water, with only color degradation, but clear foreground and background. The second row represents turbid water, with both color degradation and severe occlusion. As seen in
Fig.~\ref{fig:AblationStudyOfCRB}, the DehazeFormer Block focuses on the foreground, and its performance decreases significantly when most of the foreground is occluded. In contrast, CRB can focus on global color correction, and its heatmap more accurately reflects the distribution of color bias. This shows that the DehazeFormer Block and CRB handle different subtasks in the UIE task, with CRB assisting the DehazeFormer Block in feature extraction while performing color restoration.
Furthermore, to demonstrate the higher performance brought about by CRB, we replaced MTDA in ~\cite{Restormer} and MCC in~\cite{DNF} with CRB. The results are shown in Tab.~\ref{tab:AblationStudyOfCRB}. Additionally, RLN in the CRB significantly improved color restoration, and for comparison, BatchNorm and LayerNorm were included in the ablation study. To further highlight the benefits of CRB and channel self-attention, we also included popular attention mechanisms, SE Block~\cite{SENet} and CBAM~\cite{CBAM}, in the ablation experiment.

\begin{table}[h]
\centering
\scalebox{0.9}{
\begin{tabular}{c||cc}
\hline
Variants                       & SSIM↑           & PSNR↑ \\ \hline
WaterFormer                    & \textbf{ 0.917} & \textbf{ 31.18} \\
CRB→MTDA~\cite{Restormer}      & 0.910           &  30.01 \\
CRB→MCC~\cite{DNF}             & 0.912           &  30.43 \\
CRB→SE Block~\cite{SENet}      & 0.909           &  30.00 \\
CRB→CBAM~\cite{CBAM}           & 0.907           &  29.89 \\
RLN→BatchNorm                  & 0.916           &  31.05 \\
RLN→LayerNorm                  & 0.915           &  30.85 \\
\hline
\end{tabular}
}
\caption{Quantitative analysis of CRB ablation experiments. The bold font indicates the best performance.} \label{tab:AblationStudyOfCRB}
\end{table}

\begin{figure}[htb]
    \centering
   \begin{overpic}[width=\columnwidth]{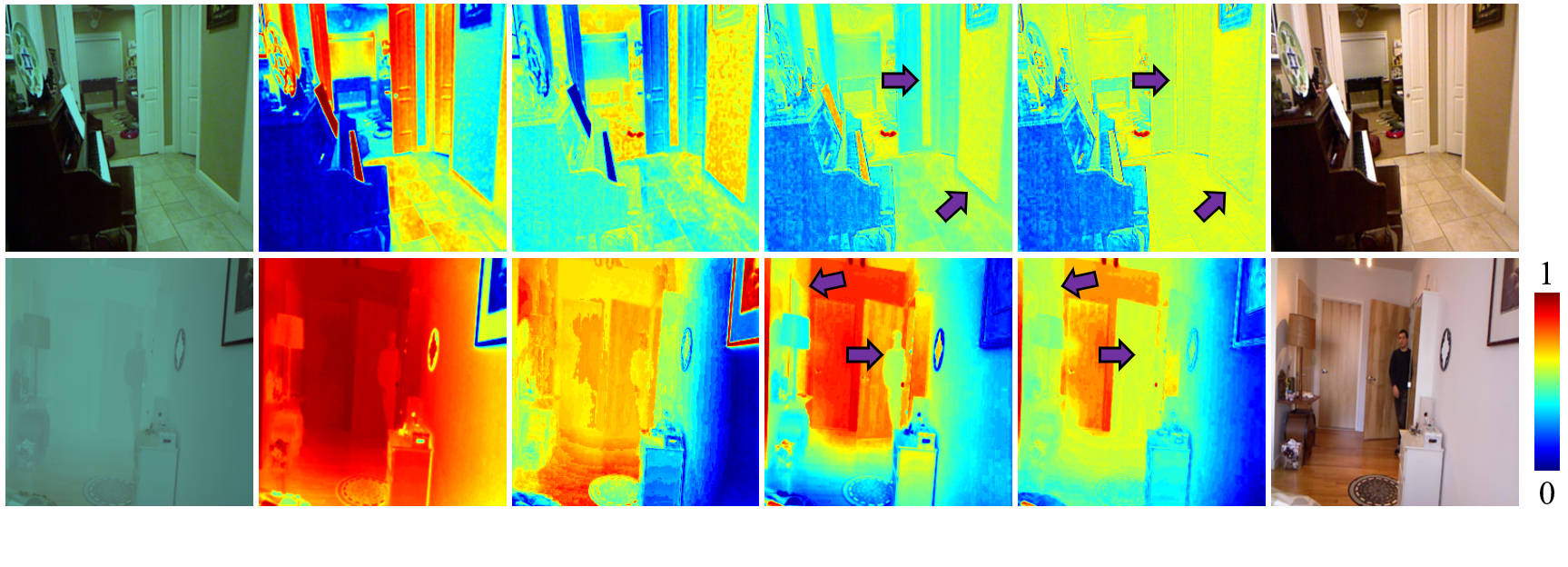} \small
    \put(6,0){(a)}
    \put(22,0){(b)}
    \put(37.5,0){(c)}
    \put(54,0){(d)}
    \put(71,0){(e)}
    \put(87,0){(f)}
   \end{overpic}
   \caption{Visualization of the CFB fusion mechanism. (a) Input, (b) result of image embedding layer, (c) result of CRB, (d) result of CFB, (e) result of addition, (f) GT.
   }\label{fig:AblationStudyOfCFB}
\end{figure}

\textbf{CFB Evaluation.} To better validate the effectiveness of the CFB, its fusion mechanism was visualized and compared with pixel-wise addition, as shown in Fig.~\ref{fig:AblationStudyOfCFB}. The first CFB example is displayed, where its two input feature maps come from the image embedding layer (composed of a convolutional layer with a 3x3 kernel size) and the 2-block CRB. We also selected clear and turbid water types for the experiment in Test-S300, which are presented in the first and second rows, respectively. In Fig.~\ref{fig:AblationStudyOfCFB}, the image embedding layer can only extract shallow features, and its heatmap reflects mainly the pixel distribution of the input. On the other hand, as described in the previous subsection, the heatmap generated by CRB displays the distribution of color bias information. From the heatmap, it can be seen that if the two feature maps are added pixel-wise, many details are lost, and the WaterFormer inference results are therefore affected. However, after fusion using the CFB, the results for both clear and turbid water are presented with more distinct layers and better detail preservation (as indicated by the purple arrow). 

\textbf{Other Evaluation.} The effectiveness of other improvements in WaterFormer was demonstrated in Tab.~\ref{tab:AblationStudyOfOtherImprovements}, including the replacement of the ReLU with the FReLU~\cite{FReLU} in the MLP structure and the replacement of the global residual reconstruction layer with the soft reconstruction layer proposed in~\cite{DehazeFormer} and optimized according to Eq.~\ref{equ:WaterFormerImageFormula}. The study on the size of the sliding window in ${{L}_{chroma}}$ can be found in Tab.~\ref{tab:AblationStudyOfWindowSize}. The study on the scaling coefficients of each loss term in the loss function can be found in Tab.~\ref{tab:AblationStudyOfLossTerm}.

\begin{table}[htb]
\centering
\scalebox{0.9}{
\begin{tabular}{c||cc}
\hline
Variants                       & SSIM↑             & PSNR↑ \\ \hline
WaterFormer                    & \textbf{ 0.917}   & \textbf{ 31.18} \\
FReLU~\cite{FReLU}→ReLU        & 0.915             & 30.82 \\
UW Soft Recon→Recon            & 0.911             & 30.26 \\
UW Soft Recon→Soft Recon~\cite{DehazeFormer} & 0.916 & 30.79 \\
\hline
\end{tabular}
}
\caption{Quantitative analysis of the ablation experiments on other improvements in WaterFormer. The bold font indicates the best performance.} \label{tab:AblationStudyOfOtherImprovements}
\end{table}

\begin{table}[htb]
\centering
\scalebox{0.95}{
\begin{tabular}{c||cc}
\hline
Sliding window size            & SSIM↑             & PSNR↑ \\ \hline
$11\times11$                   & 0.916             & 31.12 \\
$13\times13$                   & 0.916             & 31.04 \\
$15\times15$                   & \textbf{ 0.917}   & \textbf{ 31.18} \\
\hline
\end{tabular}
}
\caption{Quantitative analysis of the ablation experiments on size of the sliding window in ${{L}_{chroma}}$. The bold font indicates the best performance.} \label{tab:AblationStudyOfWindowSize}
\end{table}

\begin{table}[htb]
\centering
\scalebox{0.9}{
\begin{tabular}{ccc||cc}
\hline
${\lambda }_{1}$  & ${\lambda }_{2}$  &$ {\lambda }_{3}$  & SSIM↑             & PSNR↑ \\ \hline
1                 & 1                 & 1                 & 0.912             & 30.53 \\
2                 & 1                 & 1                 & 0.909             & 30.03 \\
1                 & 2                 & 1                 & 0.915             & 30.88 \\
1                 & 1                 & 2                 & \textbf{ 0.917}   & 31.18 \\
2                 & 2                 & 1                 & 0.912             & 30.48 \\
2                 & 1                 & 2                 & 0.912             & 30.84 \\
1                 & 2                 & 2                 & 0.916             & \textbf{ 31.29} \\
2                 & 2                 & 2                 & 0.911             & 30.20 \\
\hline
\end{tabular}
}
\caption{Quantitative analysis of the ablation experiments on scaling coefficients of each loss term in the loss function. The bold font indicates the best performance.} \label{tab:AblationStudyOfLossTerm}
\end{table}

\section{Conclusion}
\label{sec:conclusion}
In this work, we introduce WaterFormer, a versatile method to enhance the quality of underwater images. To address challenges such as haziness and color shifts, WaterFormer leverages the ViT architecture and comprises three key components: the DehazeFormer Block, the Color Restoration Block (CRB) and the Channel Fusion Block (CFB). The DehazeFormer Block transforms hazy features and extracts deep-level features. The CRB enhances colors through channel-level transpose matrices. The CFB facilitates feature fusion based on the importance of global information across different feature channels. We also incorporated the Chromatic Consistency Loss and Sobel Color Loss to ensure color fidelity and preserve fine details. Quantitative evaluations, qualitative evaluations, and ablation experiments show that our proposed network performs significantly better than various methods.

However, this study has several limitations. First, our work requires a substantial amount of paired data for training, which poses a significant challenge for practical applications. Second, we did not consider the effects of noise, such as environmental noise and camera noise, during image processing. Large-scale datasets that cover and fully consider complex types of image noise will be essential to train a more robust and generalizable model.

In our subsequent work, we will consider incorporating a denoising module into the network to reduce the impact of noise and enhance the robustness of the image correction results. In addition, we will use a contrastive learning strategy to reduce the dependency of the network on data and improve generalization performance. The denoising process will be integrated into the end-to-end network as a module instead of a separate preprocessing step. We will also explore the processing or analysis based on enhanced results, such as the underwater image classification and measures of image representability. We believe that the analysis of restored underwater images can provide incremental assistance in future underwater visual-imaging tasks.


\bibliographystyle{IEEEbib}
\bibliography{main}

\end{document}